\documentclass[sigconf, 10pt]{acmart}

\AtBeginDocument{%
  \providecommand\BibTeX{{%
    \normalfont B\kern-0.5em{\scshape i\kern-0.25em b}\kern-0.8em\TeX}}}

\setcopyright{acmcopyright}
\copyrightyear{2022}
\acmYear{2022}
\acmDOI{XXXXXXX.XXXXXXX}

\acmConference[MobiArch]{Workshop on Mobility in the Evolving Internet Architecture}{21st Oct,
  2022}{Sydney, NSW, Australia}
\acmPrice{15.00}
\acmISBN{978-1-4503-XXXX-X/18/06}

\begin{document}

\title{GCN-based Multi-task Representation Learning for Anomaly Detection in Attributed Networks}


\author{Venus Haghighi}
\affiliation{%
  \institution{School of Computing, Macquarie University}
  \streetaddress{1 Th{\o}rv{\"a}ld Circle}
  \city{Sydney}
  \state{NSW}
  \country{Australia}}
\email{venus.haghighi@hdr.mq.edu.au}

\author{Behnaz Soltani}
\affiliation{%
  \institution{School of Computing, Macquarie University}
  \streetaddress{1 Th{\o}rv{\"a}ld Circle}
  \city{Sydney}
  \state{NSW}
  \country{Australia}}
\email{behnaz.soltani@hdr.mq.edu.au}

\author{Adnan Mahmood}
\affiliation{%
  \institution{School of Computing, Macquarie University }
  \streetaddress{1 Th{\o}rv{\"a}ld Circle}
  \city{Sydney}
  \state{NSW}
  \country{Australia}}
\email{adnan.mahmood@mq.edu.au}

\author{Quan Z. Sheng}
\affiliation{%
  \institution{School of Computing, Macquarie University }
  \streetaddress{1 Th{\o}rv{\"a}ld Circle}
  \city{Sydney}
  \state{NSW}
  \country{Australia}}
\email{michael.sheng@mq.edu.au}

\author{Jian Yang}
\affiliation{%
  \institution{School of Computing, Macquarie University}
  \streetaddress{1 Th{\o}rv{\"a}ld Circle}
  \city{Sydney}
  \state{NSW}
  \country{Australia}}
\email{jian.yang@mq.edu.au}

\renewcommand{\shortauthors}{Haghighi, et al.}

\begin{abstract}
 Anomaly detection in attributed networks has received a considerable attention in recent years due to its 
 applications in a wide range of 
 domains 
 such as finance, network security, and medicine. Traditional approaches cannot be adopted on attributed networks' settings to solve the problem of anomaly detection. The main limitation of such approaches is that they inherently ignore the relational information between data features. With a rapid explosion in deep learning- and graph neural networks-based techniques, spotting rare objects on attributed networks has significantly stepped forward owing to the potentials of deep techniques in extracting complex relationships. 
 In this paper, we propose a 
 new architecture 
 on 
 anomaly detection. The main goal of designing such an architecture is to utilize multi-task learning which will enhance the detection performance. Multi-task learning-based anomaly detection is still in its infancy and only a few studies in the existing literature have catered to the same. We incorporate both community detection and multi-view representation learning techniques for extracting distinct and complementary information from attributed networks and subsequently fuse the captured information for achieving a better detection result. The mutual collaboration between two main components employed in this architecture, i.e., community-specific learning and multi-view representation learning, exhibits a promising solution to reach more effective results.  
\end{abstract}




\maketitle

\section{Introduction}
Anomalies are defined as abnormal instances which significantly deviate from the standard or normal patterns \cite{hawkins1980identification}. Anomaly detection is a data mining process, which aims to identify those rare and abnormal objects amongst a swarm of normal objects \cite{aggarwal2017introduction}. In the area of computer science, the research on anomaly detection dates back to the 1980s and has a wide range of applications in various domains, including but not limited to, spam detection in social networks, financial fraud detection, healthcare abuse monitoring systems, and intrusion detection in computer networks \cite{chandola2009anomaly}. As attributed networks have been widely used for modelling complex and real-world systems, anomaly detection in attributed networks has gained attention of researchers in both academia and industry as a vital research challenge. In attributed networks, nodes represent real-world objects and the edges represent their relationships. Comparing to plain networks, nodes and/or edges in attributed networks are associated with a set of attributes. Thus, attributed networks encompass both structure and attribute information, which is invaluable in the process of anomaly detection. The main research question is \emph{how to develop an effective anomaly detection process on attributed networks by taking the advantages of both structure and attribute information}. 

Anomaly detection in attributed networks aims to identify anomalous network objects (i.e., nodes, edges, or the accumulation of nodes and edges). Many techniques have investigated for anomaly detection in attributed networks~\cite{akoglu2015graph}. The first group of approaches rely on statistical models, which have natural limitations for capturing complex relationships. As a result, this group of techniques are unable to detect new types of anomalies and cannot easily scale for large networks \cite{akoglu2010oddball,li2014probabilistic}. 
Secondly, traditional machine learning techniques such as CUR matrix factorization \cite{mahoney2009cur} and Support Vector Machines (SVM) \cite{li2017radar}  have been proposed for the purpose of anomaly detection in networks. 
This 
group of techniques cannot be easily scaled for large-scale data efficiently. Furthermore, they suffer from high computational overhead in both execution time and memory. The aforementioned techniques are also not capable of capturing the non-linear properties of real-world objects in attributes networks. Hence, practically, the objects’ representations learned via them are not rich enough to support the process of anomaly detection. To address these challenges, recent efforts have been made to utilize the potential of deep learning techniques for the purpose of detecting anomalous objects in attributed networks \cite{ma2021comprehensive}. More recent research has been conducted for adopting Graph Neural Networks (GNNs) in the process of anomaly detection in attributed networks.
First introduced in \cite{gori2005new}, 
GNNs are powerful tools for capturing the complex interactions of structure and attributes in networks and have achieved great success amongst existing approaches. Among all proposed architectures, 
one of the most outstanding type is the Graph Convolutional Networks (GCNs) \cite{kipf2016semi}. 
The main idea of GCNs is aggregating local features from local graph neighbourhood to generate a new representation for a given node. Graph Attention Networks (GATs) allocate different weights to different nodes in a local neighbourhood \cite{velivckovic2017graph}. 

Anomalous nodes can be divided into three types: {\em global}, {\em structural}, and {\em community} anomalies. 
Global anomalies are nodes that have attributes significantly different from all other nodes in a network. Structural anomalies are those nodes that create unusual connections with other nodes (e.g., nodes forming dense links with other nodes or having connections with different communities). Community anomalies are considered as nodes that have different attributes compared to other nodes in the same community. In some literature, community anomalies are also named as local anomalies \cite{ma2021comprehensive}. 

\begin{figure}
    \centering
    \includegraphics[trim=35mm 5mm 35mm -4mm,width=1\linewidth]{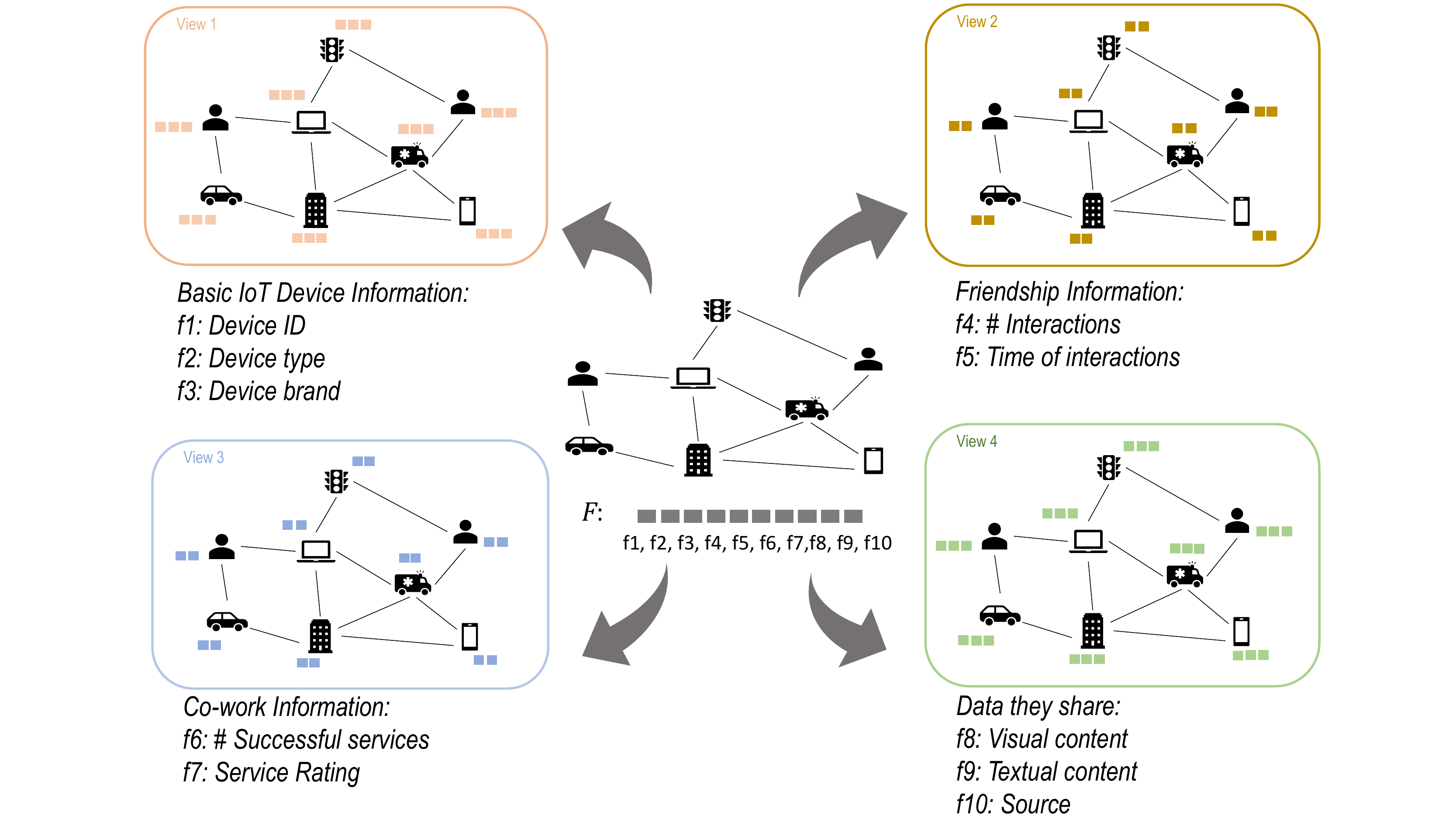}
    \caption{Different types of views for Social Internet of Things networks}
    \label{fig:img1}
\end{figure}

In this paper, we investigate the problem of encoding attributed networks under multiple views to a new shared space for extracting useful information. As different views usually correspond to different patterns and exhibit their own unique statistical features, multi-view representations provide distinct but complementary information for extracting more effective nodes' embedding in attributed networks. 
Figure~\ref{fig:img1} illustrates different types of views, which can be utilized to define a Social Internet of Things (SIoT) network such as a smart city. The first view can encompass basic device information like device ID, device type, and device brand. Second view could include friendship expressed in terms of the the number of interactions and the time of interactions between a specific device and its connected nodes. The third view may potentially contain co-work relationship between two devices in a SIoT network delineated the number of successful services when two nodes have in common in order to deliver to a third node. The service rating is another feature in the third view which is measured by the nodes receiving a specific service. The fourth view can consider data shared through out the network \cite{aljubairy2020siotpredict,sagar2022understanding}. Most of the existing studies have neglected the inherent multi-view properties of data, and treated all data features equally by concatenating them in a shared single space. In this framework, multi-view representation learning is one of the prominent modules for improving the process of anomaly detection.

We also incorporate community-specific representation learning module in the proposed framework to reduce the over-smoothing problem of convolutional operation of GNNs. Over-smoothing can lead to sub-optimal results and make nodes' representations more similar, 
making the recognition of abnormal nodes less effective \cite{wu2020comprehensive}.

The rest of the paper is organized as follows. 
Section \ref{sec:relatedwork} briefly introduces the relevant work in the literature. 
Section~\ref{sec:problem} describes our problem and its related definitions. The new proposed framework is discussed in Section \ref{sec:framework}, and finally, Section \ref{sec:conclusion} offers some concluding remarks.

\vspace{-0.063cm}
\section{Related Work}
\label{sec:relatedwork}
 A quick glimpse of literature reveals that anomaly detection has been primarily addressed in two sorts of networks, i.e., plain networks and attributed networks. In plain networks, the only available information is of the network structure such as node degree. Therefore, by utilizing some structural information, anomalous nodes can be recognized. OddBall \cite{akoglu2010oddball} detects anomalies based on the egonet patterns in plain networks. In addition, \cite{ding2012intrusion} introduces a fundamental concept of intrusion, i.e., entering a community which one does not belong to, for detecting malicious network sources. In this method, a community detection algorithm alongside a recursive search and pruning strategy have been applied for finding cut-vertices which have the highest probability of being classified as anomalies. Furthermore, as attributed networks have been increasingly promoted due to their power in modeling real-world information systems, many anomaly detection methods have been proposed in this area. 
 CODA \cite{gao2010community} 
 is a probabilistic model to detect community anomalies in attributed networks. Some other studies adopted residual analysis to spot anomalous nodes \cite{li2017radar,peng2018anomalous}. Anomalous \cite{peng2018anomalous} 
 combines 
 residual analysis and CUR decomposition simultaneously for the purpose of anomaly detection in attributed networks. 

In recent years, some GNN-based methods have been proposed for spotting anomalies in attributed networks. DOMINANT \cite{ding2019deep} is one of the first studies in this area, wherein the authors leveraged GCN for nodes embedding representations by aggregating neighbors' attributes information for the learning process. Subsequently, the reconstruction errors of nodes' representations are utilized for ranking anomalous nodes in attributed networks. COLA \cite{liu2021anomaly} 
is 
a contrastive self-supervised learning framework based on GNN for anomaly detection in attributed networks. This framework fully exploits the local information by sampling a new type of contrastive instance pair that ascertains relationships between each node and its neighbors. 
Some other related works designed GNN-based models to mitigate the over-smoothing of convolutional process in GNNs. 
ResGCN \cite{pei2022resgcn} 
applies 
the residual-based mechanism and prevents over-smoothing problem during the process of nodes' representations. ComGA \cite{luo2022comga} 
is 
a new community-aware deep GCN model for avoiding the problem of over-smoothing in the convolutional process, thereby making anomalous nodes more distinguishable. ALARM \cite{peng2020deep}
aims 
at improving the performance of anomaly detection process by considering the inherent multi-view property of data in attribute space. The said architecture supports both self-learning and user-guided learning modes which enables users to interfere in the process of anomaly detection based on their needs. 

Our proposed framework, taking the advantage of both multi-view representation learning and community-specific representation learning, aims 
at 
achieving more effective detection results. We describe the proposed framework in detail in the following sections. 

\section{Problem Statement}
\label{sec:problem}

In this section, we introduce some definitions about attributed networks, multi-view attributed networks, and modularity matrix. Then, the research problem of this paper will be introduced.
 
\vspace{3mm}
\noindent{\bfseries Definition 1 (Attributed Networks). } A static attributed network 
\begin{math}
 G=(V,E,X)
\end{math}
consists of a node set 
\begin{math}
V=\{v_1,v_2,…,v_n\}
\end{math}
where 
\begin{math}
|V|=n
\end{math}
and an edge set 
\begin{math}
 E
\end{math}
where 
\begin{math}
|E|=m
\end{math}.
The nodes attributes matrix $X\in R^{n\times d}$, here the $i-th$ row vector $x_i$ is the attributes value for the nodes $v_i$.

\vspace{1mm}
\noindent{\bfseries Definition 2 (Multi-View Attributed Networks). }In multi-view attributed networks, each node $v_i$ is associated with a set of $d-dimensional$ features $F={f_1,f_2,...,f_d}$ which can be represented by $K$ distinct views or feature spaces.

\vspace{1mm}
\noindent{\bfseries Definition 3 (Modularity Matrix).  }Modularity is a metric which represents arrangement of edges statistically. It quantifies the number of edges falling within group minus the expected number in as equivalent network with edges placed at random. The modularity value indicates the possible presence of community structure. Modularity matrix $B=[b_{ij}] $ of network can be defined as: $b_{ij}=a_{ij}-k_ik_j/2m$ where $a_{ij}$ is the probability of existing edges between nodes $v_i$ and $v_j$ belonging to the same community, $k_ik_j/2m$ is the expected number of edges between nodes $v_i$ and $v_j$ which are placed randomly, $k_i$ is the degree of node $v_i$, and $m=\frac{1}{2}\sum_ik_i$ is the total number of edges\cite{newman2006modularity}.

\vspace{1mm}
\noindent{\bfseries Research Problem.   }Given a multi-view attributed network $G$ represented by adjacency matrix $A$ and K-view attributes $X=(X^1,X^2,...,X^k)$, where $X^i$ is the attribute vector corresponding to the $i^{th}$ view with the dimensions of $D_i$. the problem aims to detect nodes that differ significantly from the majority nodes from the perspectives of structure and attributes.

\section{The Proposed Framework}
\label{sec:framework}

\begin{figure*}
    \centering
    \includegraphics[width=1\textwidth]{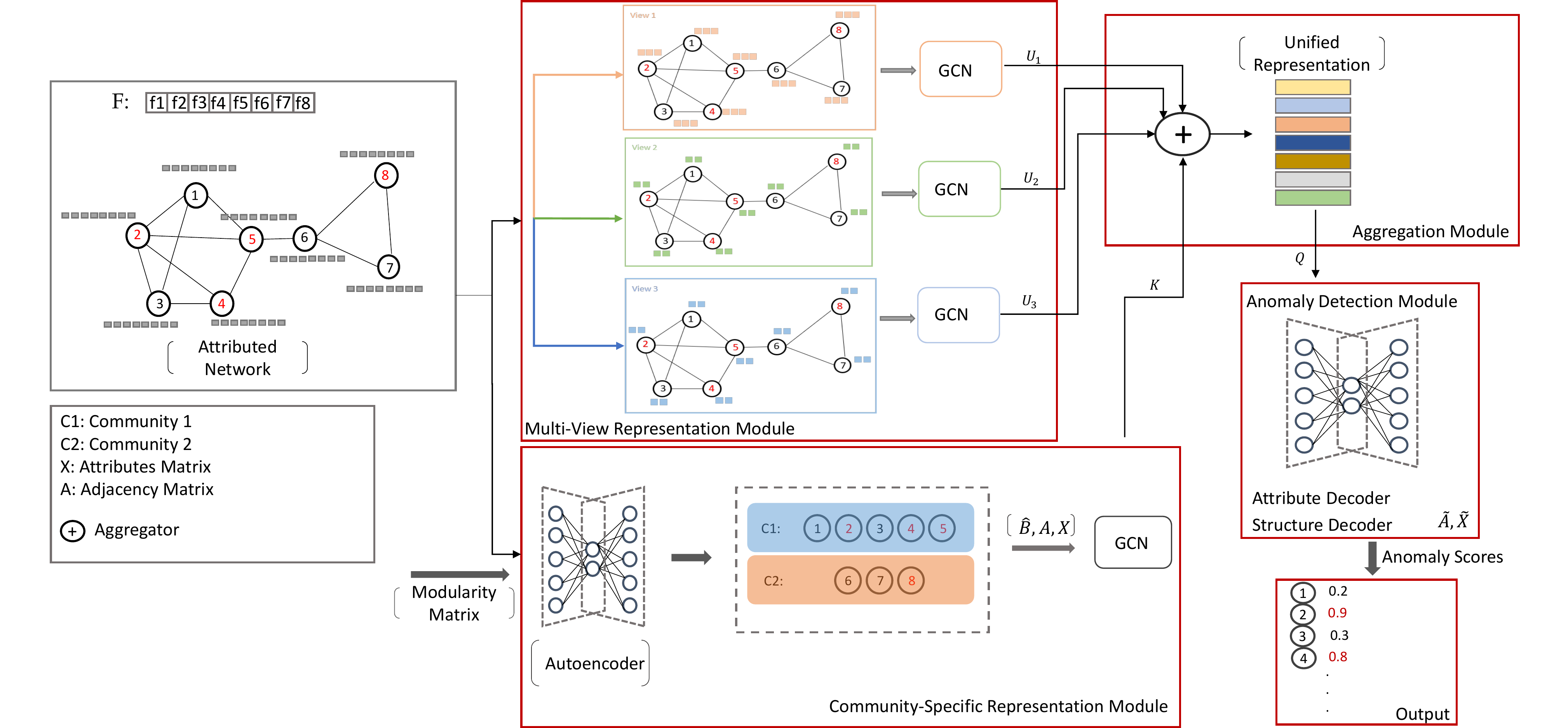}
    \caption{Multi-task learning framework for anomaly detection on attributed networks. Nodes marked by red color indicate anomalies in the given network.}
    \label{fig:img2}
\end{figure*}

In this section, we illustrate the new proposed multi-task learning architecture in 
Figure~\ref{fig:img2}. The entire framework includes four main components. First, multi-view representation learning component 
adopts 
multiple graph neural networks (GNNs) for the purpose of embedding the given multi-view attributed network. Second, community-specific representation learning module 
aims at capturing community structure of the given attributed network. Community-specific representation module leverages both autoencoder and graph neural network (GNN) to encode the structure of the given attributed network. Third component of the proposed framework is aggregator. Aggregating multi-task representation learned from the first and second components 
is 
handled in this module in order to reach a unified representation across all tasks. After the training process, the anomalies are ranked in the last component which is named anomaly detection module. The ranking process 
is 
done based on the calculated joint reconstruction errors from both perspectives of attributes and structure. The larger the reconstruction error have, the more likely the nodes are to be anomalous on the network. In the next sub-sections, each of the proposed components will be introduced in details.  

\subsection{Multi-View Representation Module}

The multi-view representation learning module consists of multiple independent GNNs to learn knowledge from each view separately. We adopt deep graph convolutional networks (GCNs), which is one of the most prominent models in GNNs.  Each view is fed into a separate GCN for the learning process. The core of GCN 
updates the feature vector $h_i$ which 
corresponds 
to each node $v_i$. The updating process 
is then 
calculated via aggregating features of its neighboring nodes $v_j$:
\begin{equation}
\ h_i^{(l)}=f_{relu} \left(\sum_{{v_j\in{v_i}\cup N(v_i)}} a_{ij}W^{(l)}h_j^{(l-1)}\right),
\end{equation}
where $h_i^{(l)}$ is the learned representation corresponding to node $v_i$ at the $l^{th}$ layer, $W^{(l)}$ is trainable weight matrix at the $l^{th}$ layer, $f_{relu}(x)=max(0,x)$ is a non-linear activation function, $a_{ij}$ is the $(i,j)^{th}$ entry of $\hat{A}$ which is calculated by $(\hat{A}=D^{-\frac{1}{2}}\bar{A}D^{-\frac{1}{2}}, \bar{A}=A+I_n, D_{ii}=\sum_j \bar{A_{ij}})$, and $h_i^{(0)}$ is the original attribute vector of node $v_i$. The non-linear activation function $f_{relu}$ can capture the non-linear relationships between data by considering both attribute information $X$ and graph structure $A$. As a result of convolution operations, each of these GCNs produces a learned high-quality embedding representation $U_i$ from different views.

\subsection{Community-Specific Representation Module}
We utilize a deep autoencoder architecture to capture the community structure of graphs by avoiding the problem of linear mapping process of traditional community detection methods. According to the constructed modularity matrix $B$ of the graph, the learning process is initiated. In fact, The modularity matrix represents the division of nodes in different communities. The autoencoder consists of two main components: the first component is encoder, and the second one is decoder. The community-specific learned representation corresponding to $l^{th}$ layer is defined as \cite{luo2022comga}:
\begin{equation}
\ H^l=f_{relu} \left(W^lH^{l-1}+b^l\right),
\end{equation}
where $f_{relu}$ is the activation function, $W^l$ and $b^l$ represent the trainable weight matrix and bias of the $l^{th}$ iteration in the encoder, respectively.  The input of encoder $H^0$ is the modularity matrix $B$.

In the decoder part, we 
reconstruct the original data from the learned representation of the encoder:
\begin{equation}
H'^l=f_{relu}\left(W^lH'^{l-1}+b^l \right)
\end{equation}
The reconstruction loss function of the autoencoder is presented as follows:
\begin{equation}
\ L_{res}=\| B-\hat{B}\|_F^2  ,
\end{equation}
where $\hat{B}$ is the estimated modularity matrix by decoder, and $\|B-\hat{B}\|_F$ is the Frobenius norm.The next step is fusing the learned modularity matix $\hat{B}$ into a GCN model for encoding the attributed network based on the learned modularity matrix $\hat{B}$, attributed matrix $X$, and structure information $A$. Thus, the network representation of the $l^{th}$ layer in GCN mode, $Z_l$, can be learned by the following equation:
\begin{equation}
\ Z^l=f_{relu}\left(D^{-\frac{1}{2}}\bar{A}D^{-\frac{1}{2}}W^{l-1}Z^{l-1}\right) ,
\end{equation}
where $\bar{A}=A+I_n$, $D_{ii}=\sum_j \bar{A_{ij}}$, $I$ is the identity diagonal matrix of the adjacency matrix $A$, and $W$ is the weight matrix of graph convolutional network.

Therefore, the community-specific  module takes community structure information (modularity matrix $B$), node attributes ($X$), and network structure ($A$) as its input, and then 
extracts more effective node embedding representations by combining the two learned representations $Z^l$ and $H^l$ as follows:
\begin{equation}
\ K=\left(Z^{l}\|H^{l}\right) .
\end{equation}

\subsection{Aggregation Module}
Aggregation module aims at bringing all low-dimensional representations of both multi-view representations component and community-specific representations component in a shared space. It is significant to put emphasize on the fact that each of these learned representations provides us with distinct and complementary description of data on the given attributed network.Therefore, taking the advantage of complementary information between different representations to aggregate multiple representations is more sensible. In this section, we consider two kinds of aggregation strategies which are described in the following.

\vspace{2mm}
\noindent\textbf{Concatenation:} All learned representations of both multi-view representations module and community-specific representation module are concatenated into a new shared vector space. 
This solution is one of the traditional strategies for combining all of learned representations together.

\vspace{2mm}
\noindent\textbf{Weighted Aggregation:} In this strategy, different weights are given to the different learned representations from both modules, and then summing up them. In fact, each of these learned representations has a distinct proportion in the final representation. Hence, the final result $Q$ is described as the following:
\begin{equation}
\small
\ Q= \left\{Q|Q= \sum_{i=1}^k\alpha_iU_i+\beta K, \alpha_i \geq 0, \beta \geq 0, \sum_{i=1}^k \alpha_i + \beta=1 \right\}
\end{equation}

\subsection{Anomaly Detection Module}
The output of aggregation module is a combined encoded representation $Q$ from multiple perspectives. In this section, we focus on the reconstruction of the the original attributed network from the combined encoded representation. The corresponding reconstruction error can be considered as a measure for anomaly detection. Anomaly detection module consists of two main parts, namely 1) Structure Decoder, and 2) Attribute Decoder.

\vspace{2mm}
\noindent\textbf{Structure Decoder:} The structure decoder takes the latent representation $Q$ as input and then decodes it for the purpose of reconstruction of the original adjacency matrix which is named $\tilde A $. Anomalous nodes are determined from the perspective of structure by the calculation of inner product between them as follows:
\begin{equation}
\ \tilde{A}=sigmoid \left(QQ^T\right).
\end{equation}

Structure reconstruction error can be an indicator for detecting anomalous nodes on the given attributed network. The structure of a specific node can be well reconstructed through the structure decoder if the node has the low probability of being 
anomalous. 
In other word, anomalous nodes can not be fully reconstructed because they do not comply with the majority of data.

\vspace{2mm}
\noindent\textbf{Attribute Decoder:} For determining the anomalous nodes from the perspective of attributes, we utilize a fully-connected graph convolutional network to take the learned latent representation $Q$ as input and decode it to generate reconstructed attribute information as follows:
\begin{equation}
\ \tilde{X}=f_{relu} \left( ZW+B \right),
\end{equation}
where $W$ is a trainable weighted matrix, and $B$ is the bias term. The combined reconstruction error from the both perspectives of structure and attribute can be defined as follows:
\begin{equation}
\ L=L_s+L_a  ,
\end{equation}
where $L_s$ is structure loss function, $L_a$ is attribute loss function, and $L$ is total loss function which can be defined as sum of structure and attribute loss functions. For measuring the quality of reconstruction structure and attribute information, Frobenius norm can be used as follows:
\begin{equation}
\ L=(1-\lambda)||A-\tilde{A}||_F^2+\lambda||X-\tilde{X}||_F^2  ,
\end{equation}
where $\lambda$ is the balance parameter between the structure and attributes reconstructions. The anomaly detection module aims at minimizing the above defined joint loss function to measure the score of anomalous nodes on the given attributed network.

\section{Conclusion}
\label{sec:conclusion}
In this paper, we propose a novel deep framework that supports multi-task learning for improving the performance of anomaly detection in attributed networks. The main idea behind our framework is to utilize multi-task learning that contains two main components, i.e., multi-view representation learning and community-specific representation learning. Subsequently, aggregation of the proposed components contributes to more effective low-dimensional nodes' representations and reveals more anomalous nodes. Based on the theoretical results, our model can be regarded as a potential approach for enhancing the performance of anomaly detection process in attributed networks. In the future, we will validate our proposed framework via experimental studies using both synthetic and real-world datasets.

\bibliographystyle{unsrt}
\bibliography{sample-base}


\end{document}